\documentclass[letterpaper]{article} 
\usepackage{aaai2026}  
\usepackage{times}  
\usepackage{helvet}  
\usepackage{courier}  
\usepackage[hyphens]{url}  
\usepackage{graphicx} 
\usepackage{natbib}  
\usepackage{caption} 
\usepackage{amsmath}
\usepackage{amssymb}
\usepackage{mathtools}
\usepackage{amsthm}
\usepackage[capitalize,noabbrev]{cleveref}

\usepackage{algorithm}
\usepackage{algorithmic}

\theoremstyle{plain}

\newcommand{\SMILES}{\textsc{SMILES}}
\newcommand{\R}{\mathbb{R}}
\newcommand{\softmax}{\mathrm{softmax}}

\usepackage{multirow}
\usepackage{microtype}
\usepackage{graphicx}
\usepackage{subcaption}
\usepackage{booktabs}

\usepackage{newfloat}
\usepackage{listings}
\DeclareCaptionStyle{ruled}{labelfont=normalfont,labelsep=colon,strut=off} 
\floatstyle{ruled}
\newfloat{listing}{tb}{lst}{}
\floatname{listing}{Listing}
\nocopyright

\title{Template-Free Retrosynthesis with Graph-Prior Augmented Transformers}
\author{
Youjun Zhao
}
\affiliations{
City University of Hong Kong
}

\usepackage{bibentry}

\begin{document}

\maketitle

\begin{abstract}
Retrosynthesis reaction prediction aims to infer plausible reactant molecules for a given product and is a important problem in computer-aided organic synthesis.
Despite recent progress, many existing models still fall short of the accuracy and robustness required for practical deployment.
In this paper, we present a template-free, Transformer-based framework that removes the need for handcrafted reaction templates or additional chemical rule engines.
Our model injects molecular graph information into the attention mechanism to jointly exploit SMILES sequences and structural cues, and further applies a paired data augmentation strategy to enhance training diversity and scale.
Extensive experiments on the USPTO-50K benchmark demonstrate that our approach achieves state-of-the-art performance among template-free methods and substantially outperforms a vanilla Transformer baseline.
\end{abstract}

\section{Introduction}
Retrosynthesis aims to infer plausible reactant sets or reaction routes for a target product molecule and is a core problem in computer-aided organic synthesis.
It is particularly important in drug discovery, where one of the major bottlenecks is to efficiently synthesize novel and structurally complex compounds.
The underlying search space is enormous---millions of compounds and reactions have been reported---and a single product typically admits multiple valid disconnection strategies, making purely manual retrosynthesis design difficult and time-consuming.

Recent work has proposed a variety of algorithms to assist and automate retrosynthesis.
AI-driven approaches span both \emph{single-step} and \emph{multi-step} settings.
Although often studied separately, these two levels are tightly coupled: stronger single-step predictors naturally improve multi-step search success rates and reduce search time, while multi-step planning introduces additional evaluation criteria and constraints that can in turn guide the design of better single-step models.
Within the single-step setting, methods are further divided into \emph{selection-based} and \emph{generation-based} approaches depending on whether they enumerate candidate reactants from a fixed set or directly generate full reactant structures.

A common way to categorize existing methods is by their use of reaction templates.
Template-based methods encode expert-defined or automatically extracted reaction rules and typically achieve high accuracy when test reactions resemble templates in the library, but cannot propose reactions beyond it.
Template-free methods directly predict reactants from products using learned models without explicit templates, offering greater flexibility and the potential to generate novel transformations, but they face challenges in achieving high accuracy and ensuring chemical validity.
Semi-template methods combine both ideas, usually via reaction-center identification and synthon-based generation, and aim to balance template coverage, diversity, and interpretability.

In this work, we focus on \emph{single-step, template-free} retrosynthesis.
We study a Transformer-based model that (i) injects molecular graph priors into multi-head attention via a Gaussian-style distance prior and atom mapping, (ii) employs a data augmentation strategy that performs both representation-level and data-scale augmentation on paired \SMILES\ strings, and (iii) does not rely on additional reaction templates or domain-specific rule engines.

Our main contributions are:
\begin{itemize}
  \item We propose a Transformer-based retrosynthesis architecture that combines \SMILES\ sequence information with molecular graph information, enabling the model to exploit both sequential and structural properties of molecules.
  \item We design a paired data augmentation strategy that enhances molecular representations and enlarges the training set by enumerating different \SMILES\ roots and reordering product--reactant pairs, which leads to significantly improved performance and generalization.
  \item We conduct experiments on the standard USPTO-50K benchmark. Our method surpasses existing template-free methods and approaches, and even exceeds representative template-based and semi-template baselines under the same evaluation protocol.
\end{itemize}

\section{Related Work}
\subsection{Template-Based Methods}
Template-based methods depend on reaction template databases that encode core reaction rules.
Templates are typically specified by experts or extracted from reaction corpora, and can be interpreted as symbolic representations of local reaction patterns.
Representative works include NeuralSym~\cite{neural-symbolic}, RetroSim~\cite{computer-assisted}, GLN~\cite{Retrosynthesisprediction} and LocalRetro~\cite{Deepretrosynthetic}, which study reaction template scoring and application.
Such methods often achieve strong accuracy but cannot generate reactions outside the template space and are limited in diversity.

\subsection{Template-Free Methods}
Template-free methods do not rely on explicit reaction templates or additional chemical knowledge and directly learn to map products to reactants.
Molecules can be represented as \SMILES\ sequences or molecular graphs, leading to sequence-based and graph-based template-free models.
Early work~\cite{Retrosyntheticreaction} used sequence-to-sequence models with BiLSTM encoders and decoders to predict reactant sequences from product sequences.
Following the success of Transformers~\cite{vaswani2017attention} in machine translation and other NLP tasks, many studies~\cite{Automaticretrosynthetic, Learningtomake,Predictingretrosynthetic,State-of-the-artaugmented} treat retrosynthesis as a machine translation problem and adopt Transformers as backbones.
Karpov et al.\cite{Atransformermodel} first used a pure Transformer sequence model for retrosynthesis.
GTA~\cite{GTAGraph} investigated the untapped potential of sequence-based models by injecting graph information into the Transformer architecture, while Graph2SMILES~\cite{Permutationinvariant} replaces the sequence encoder with a graph encoder to ensure permutation invariance and robustness to \SMILES\ reordering.
Template-free methods usually produce more diverse and novel reactions, but may yield invalid molecules and struggle to reach the top accuracy of template-based approaches.

\subsection{Semi-Template Methods}
Semi-template methods combine the strengths of template-based and template-free methods.
Most existing approaches~\cite{retroprime,retroxpert,Agraphtographs,Learninggraph} first identify reaction centers or breaking bonds and then transform the product into intermediate synthons using RDKit\cite{rdkit}, followed by reactant generation from synthons via selection\cite{Learninggraph}, graph generation~\cite{Agraphtographs}, or \SMILES\ generation~\cite{retroprime,retroxpert}.
GraphRetro~\cite{Learninggraph}, for instance, identifies reaction centers on the product, attaches leaving groups, and then selects or generates reactants.
These methods are often competitive in accuracy and provide better interpretability, but still rely on chemistry tools and templates.

\begin{figure}[t]
  \centering
  \includegraphics[width=0.48\textwidth]{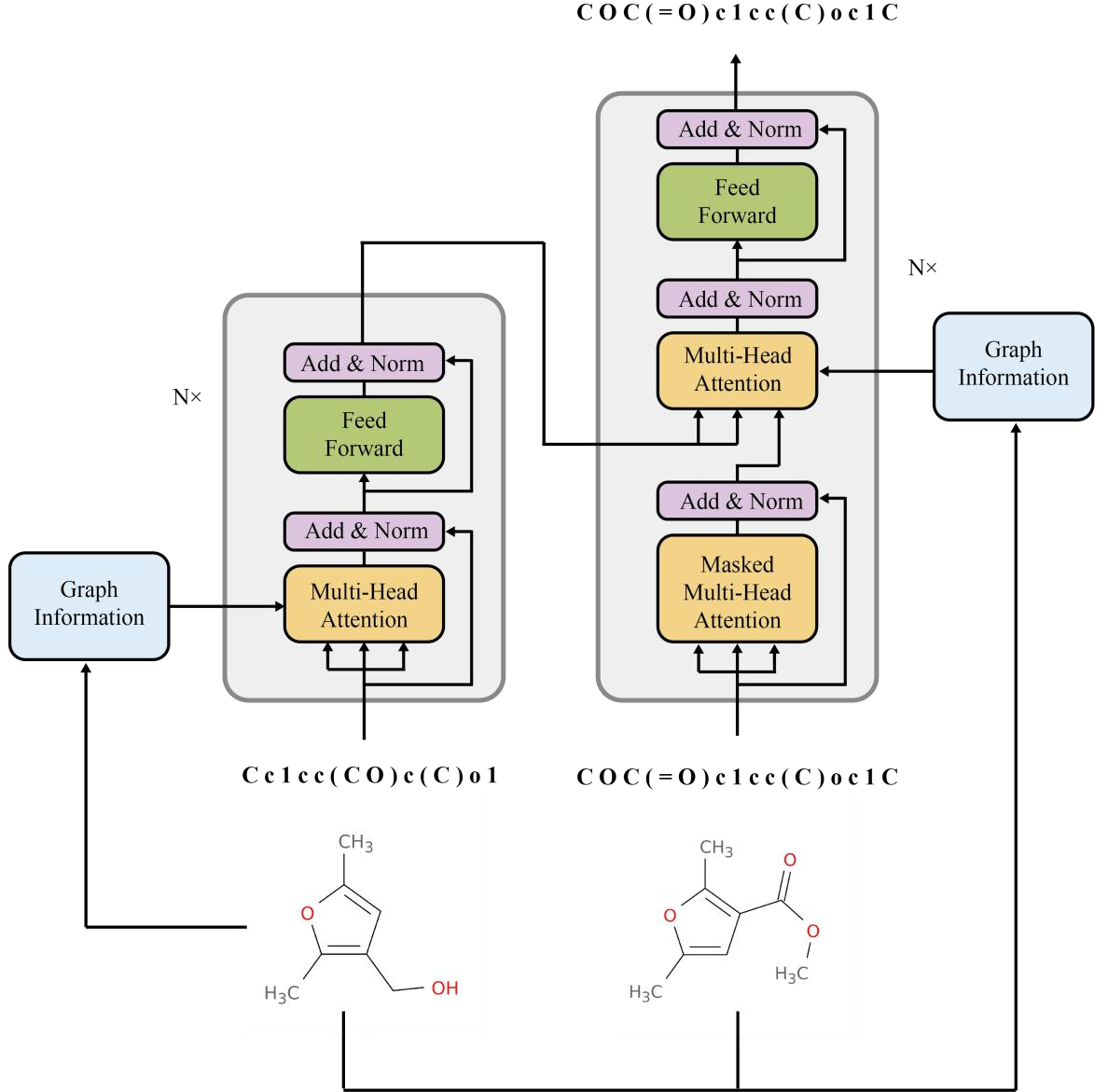}
  \caption{Overall architecture of the proposed model.
  Molecular graph information is injected into multi-head attention as structural priors, and paired \SMILES\ augmentation generates multiple product--reactant pairs for training.}
  \label{fig:framework}
\end{figure}

\section{Method}
In this section we describe the proposed Transformer-based retrosynthesis model.
\cref{fig:framework} illustrates the overall architecture.
Our method builds on a standard encoder--decoder Transformer and injects molecular graph information into multi-head attention.
Based on the one-to-one correspondence between atoms in \SMILES\ and nodes in the molecular graph, the attention module is guided to focus more on chemically relevant tokens with truncated attention links.

\subsection{Overview and Notation}
Let the product molecule be represented by a tokenized \SMILES\ sequence
$x = (x_1,\dots,x_{T_x})$ and reactants by $y = (y_1,\dots,y_{T_y})$.
We model the conditional probability of reactants given the product as
\begin{equation}
p_\theta(y\mid x)
= \prod_{t=1}^{T_y} p_\theta(y_t \mid y_{<t}, x),
\label{eq:autoregressive}
\end{equation}
where $\theta$ denotes all parameters.
The model is trained by minimizing the negative log-likelihood over the training set.

\subsection{Molecular \SMILES\ Sequences}
\SMILES\ encodes a molecule as a linear string.
Although multiple valid \SMILES\ strings can represent the same molecule, cheminformatics tools such as RDKit can generate a canonical \SMILES\.
Following standard practice, we tokenize atom symbols and non-atom tokens separately, including parentheses and ring indices.
This tokenization allows us to map atom tokens back to graph nodes and to treat syntax tokens (e.g., branch and ring markers) explicitly in the sequence model.

\subsection{Transformer Architecture}
We adopt the encoder--decoder Transformer architecture proposed by Vaswani et al.
The encoder maps the input product sequence to contextual representations.
The decoder autoregressively generates the reactant sequence conditioned on encoder outputs.

The core component is multi-head self-attention.
Given query, key and value matrices $Q,K,V \in \R^{T\times d}$, a single-head attention computes:
\begin{equation}
S = \frac{QK^\top}{\sqrt{d_k}},
\end{equation}
\begin{equation}
\mathrm{Attn}(Q,K,V) = \softmax(\mathrm{Mask}(S,M)) V,
\end{equation}
where $M\in\{0,1\}^{T\times T}$ is a binary mask.
For decoder self-attention, $M$ enforces causal masking so that each position can only attend to previous tokens; for encoder self-attention, $M$ may encode padding or other structural constraints.

Multi-head attention repeats the above operation with multiple sets of $(W_Q,W_K,W_V)$ and concatenates their outputs, followed by a feed-forward network and residual connections.
Compared with RNNs such as LSTMs, the Transformer can better capture long-range dependencies and is easier to train with parallel computation, which is beneficial for long \SMILES\ strings.

\subsection{Representation of Intra-Molecular Graph Information}
We view the product molecule as a graph $G=(V,E)$ where nodes are atoms and edges are chemical bonds.
Let $D_{ij}$ denote the shortest path distance between atom $i$ and atom $j$ in the graph.
To encode graph structure, we construct distance-specific binary matrices:
\begin{equation}
m^{(d)}_{ij} =
\begin{cases}
1, & D_{ij} = d,\\
0, & D_{ij} \neq d,
\end{cases}
\qquad d\in\{1,2,3,4\}.
\end{equation}
These matrices indicate whether two atoms are $d$-hop neighbors.
Since our attention operates on \SMILES\ tokens, we align atom tokens with graph nodes and propagate these masks to the token level.

We then form a graph-informed bias matrix $B_{\text{intra}}$ (either a weighted combination of $m^{(d)}$ or a soft kernel) and add it to the self-attention logits:
\begin{equation}
\widetilde{S}
= S + \lambda_{\text{intra}} B_{\text{intra}},
\end{equation}
where $\lambda_{\text{intra}}$ controls the strength of the prior.
Intuitively, this encourages attention to focus on chemically nearby atoms.

To connect with the Gaussian prior description in the original thesis, one can define a soft distance prior:
\begin{equation}
g_{ij} = \exp\!\bigg(- \frac{D_{ij}^2}{2\sigma^2}\bigg),
\end{equation}
and set $B_{\text{intra}} = g$.
In practice, both hard and soft variants follow the same additive-bias form and do not change the overall model structure.

\subsection{Representation of Inter-Molecular Cross Graph Information}
In addition to intra-molecular structure, we exploit cross-graph relations between reactants and products.
During many reactions, only a small part of the molecule changes while most substructures are preserved.
Atom mapping tools can identify correspondences between atoms in the product and reactants.

Suppose we obtain an atom mapping between product atoms and reactant atoms.
We build a binary alignment matrix $B_{\text{cross}} \in\{0,1\}^{T_x \times T_y}$ at the token level:
\begin{equation}
(B_{\text{cross}})_{ij} =
\begin{cases}
1, & \text{if $(i,j)$ is a mapped atom pair},\\
0, & \text{otherwise}.
\end{cases}
\end{equation}
This matrix is used as a bias for encoder--decoder (cross) attention:
\begin{equation}
\widetilde{S}^{\text{cross}}
= \frac{Q_{\text{dec}} K_{\text{enc}}^\top}{\sqrt{d_k}}
+ \lambda_{\text{cross}} B_{\text{cross}},
\end{equation}
and the attention is computed via $\softmax(\widetilde{S}^{\text{cross}})$.
Thus the decoder is encouraged to attend to aligned regions that correspond to unchanged structural motifs.

\subsection{Data Augmentation}
We introduce two data augmentation strategies as below:
\begin{itemize}
  \item \textbf{Representation augmentation}.
  For each product--reactant pair, we randomly choose an atom as the \SMILES\ root for the product, regenerate the product \SMILES\ based on the new traversal, and reorder the reactant \SMILES\ accordingly using atom mapping.
  We remove irrelevant symbols to obtain a new consistent product--reactant pair.
  This produces multiple equivalent training examples with different linearizations but consistent molecular graphs.
  \item \textbf{Data-scale augmentation}.
  Following prior work, we enlarge the training set by a factor of $20\times$ using the above enumeration, while keeping the validation and test sets unchanged for fair evaluation.
\end{itemize}

The training objective remains standard cross-entropy under teacher forcing:
\begin{equation}
\mathcal{L}(\theta)
= - \sum_{(x,y)\in\mathcal{D}}
\sum_{t=1}^{T_y} \log p_\theta(y_t \mid y_{<t}, x).
\end{equation}

\begin{table*}[t]
\caption{Top-$K$ accuracy (\%) of single-step retrosynthesis methods on USPTO-50K.}
\label{tab:sota}
\centering
\begin{small}
\begin{tabular}{lcccc}
\toprule
Method & Top-1 & Top-3 & Top-5 & Top-10 \\
\midrule
\multicolumn{5}{l}{\textit{Template-based}} \\
RetroSim~\cite{computer-assisted} & 37.3 & 54.7 & 63.3 & 74.1 \\
NeuralSym~\cite{neural-symbolic} & 44.4 & 60.3 & 72.4 & 78.9 \\
GLN~\cite{Retrosynthesisprediction} & 52.5 & 69.0 & 75.6 & 83.7 \\
LocalRetro~\cite{Deepretrosynthetic} & 53.4 & 77.5 & 85.9 & 92.4 \\
\midrule
\multicolumn{5}{l}{\textit{Semi-template}} \\
G2Gs~\cite{Agraphtographs} & 48.9 & 67.6 & 72.5 & 75.5 \\
GraphRetro~\cite{Learninggraph} & 53.7 & 68.3 & 72.2 & 75.5 \\
RetroXpert~\cite{retroxpert} & 50.4 & 61.1 & 62.3 & 63.4 \\
RetroPrime~\cite{retroprime} & 51.4 & 70.8 & 74.0 & 76.1 \\
R-SMILES~\cite{Root-alignedSMILES} & 49.1 & 68.4 & 75.8 & 82.2 \\
\midrule
\multicolumn{5}{l}{\textit{Template-free}} \\
BiLSTM & 37.4 & 52.4 & 57.0 & 61.7 \\
Transformer~\cite{vaswani2017attention} & 42.0 & 57.0 & 61.9 & 65.7 \\
GTA~\cite{GTAGraph} & 51.1 & 67.6 & 74.8 & 81.6 \\
Dual-TF~\cite{Towardsunderstanding} & 53.3 & 69.7 & 73.0 & 75.0 \\
MEGAN~\cite{Moleculeedit} & 48.1 & 70.7 & 78.4 & 86.1 \\
Tied Transformer~\cite{Valid} & 47.1 & 67.2 & 73.5 & 78.5 \\
Aug. Transformer~\cite{State-of-the-artaugmented} & 53.5 & --   & 81.0 & 85.7 \\
R-SMILES~\cite{Root-alignedSMILES} & 53.6 & 75.8 & 81.3 & 84.6 \\
\textbf{Ours} & \textbf{54.3} & \textbf{78.0} & \textbf{85.2} & \textbf{91.1} \\
\bottomrule
\end{tabular}
\end{small}
\end{table*}

\section{Experiments}
\subsection{Dataset}
We evaluate our approach on the widely used USPTO-50K dataset~\cite{what'swhat}, which contains 50{,}016 reactions annotated with 10 reaction classes.
We follow the same train/validation/test split as~\cite{Retrosynthesisprediction}, using 80\%/10\%/10\% of the data for training, validation, and testing, respectively.

\subsection{Data Augmentation Protocol}
We adopt a 20$\times$ data-scale augmentation on the training split.
During training, for each reaction a variety of product roots and \SMILES\ enumerations are sampled to generate multiple product--reactant pairs.
No augmentation is applied to the test set to ensure fair comparison with previous work.

\subsection{Evaluation Metrics}
We evaluate performance using top-$K$ accuracy on the test set with $K \in \{1,3,5,10\}$.
A prediction is considered correct if the ground-truth reactant set appears in the top-$K$ candidates produced by the model.

\subsection{Implementation Details}
Our implementation is based on the OpenNMT toolkit~\cite{opennmt} and PyTorch.
We use RDKit~\cite{rdkit} to construct graph distance matrices and atom mapping matrices.
The network consists of a 6-layer Transformer encoder and a 6-layer Transformer decoder with 8 attention heads.
Dropout is set to 0.3.
For relative position encoding, we use a maximum relative distance of 4.

We use the Adam optimizer with learning rate 2.0.
An early stopping strategy is adopted: training stops when the validation loss and accuracy do not improve within 40 epochs and every 1000 training steps.
All experiments are conducted on a single NVIDIA RTX 3070 GPU; one full training run takes roughly 24 hours.
We tune early stopping, dropout, depth and maximum relative distance to obtain the best performance.

\subsection{Comparison with State-of-the-Art Methods}
\cref{tab:sota} compares the proposed method with representative template-based, semi-template and template-free baselines on USPTO-50K.
All numbers follow the standard evaluation protocol from the literature.

Our model achieves 54.3/78.0/85.2/91.1\% top-1/3/5/10 accuracy.
Notably, the top-10 accuracy exceeds 90\% for the first time among template-free methods.
Compared with the strongest template-free baseline R-SMILES, our method improves top-1/3/5/10 accuracy by 0.7/2.2/3.9/6.5 percentage points, respectively.
These results demonstrate that injecting graph priors and using data augmentation substantially strengthens the template-free Transformer.

\subsection{Ablation Study}
To validate the effectiveness of each component, we perform an ablation study on three modules:
(i) graph information (including intra-molecular and cross-graph priors),
(ii) representation augmentation, and
(iii) data-scale augmentation.
\cref{tab:ablation} reports the results.

\begin{table}[t]
\caption{Ablation results on our designs.}
\label{tab:ablation}
\centering
\begin{small}
\begin{tabular}{ccc|cccc}
\toprule
Rep. & Scale & Graph & \multirow{2}{*}{Top-1} & \multirow{2}{*}{Top-3} & \multirow{2}{*}{Top-5} & \multirow{2}{*}{Top-10} \\
Aug. & Aug. & Info  &       &       &       &        \\
\midrule
 &  &  & 42.0 & 57.0 & 61.9 & 65.7 \\
 &  & $\checkmark$ & 47.3 & 67.8 & 73.8 & 80.1 \\
$\checkmark$ &  & $\checkmark$ & 50.7 & 72.8 & 80.4 & 86.0 \\
 & $\checkmark$ & $\checkmark$ & 51.5 & 67.5 & 74.5 & 81.4 \\
$\checkmark$ & $\checkmark$ &  & 53.6 & 75.8 & 81.3 & 84.6 \\
$\checkmark$ & $\checkmark$ & $\checkmark$ & \textbf{54.3} & \textbf{78.0} & \textbf{85.2} & \textbf{91.1} \\
\bottomrule
\end{tabular}
\end{small}
\end{table}

Compared with the vanilla Transformer baseline, adding graph information alone improves top-1 accuracy from 42.0\% to 47.3\%, showing that explicit structural priors help compensate for the limitations of pure \SMILES\ representations.
Combining graph priors with representation augmentation further improves performance.
When both representation and data-scale augmentation are enabled without graph information, top-1 accuracy reaches 53.6\%.
Finally, the full model that combines all three components obtains the best results, with an 11.9\% top-1 improvement over the vanilla Transformer.


\section{Discussion and Limitations}
Our experiments show that a carefully designed Transformer with graph priors and data augmentation can reach and even exceed the performance of many template-based or semi-template methods on USPTO-50K, while retaining the flexibility of template-free prediction.

However, our method still has several limitations.
First, we do not explicitly verify the chemical validity of generated reactant molecules.
The model may still produce invalid or syntactically incorrect \SMILES\ in some cases.
Second, we have not evaluated on the larger USPTO-full dataset, where atom-mapping noise is more severe and the scale is much larger.
Handling noisy mappings and large-scale data remains an open challenge.
Third, our current use of molecular graphs is still relatively simple; more sophisticated ways of combining graph and sequence representations may further improve performance.

\section{Conclusion}
We proposed a Transformer-based template-free retrosynthesis model that incorporates molecular graph priors into multi-head attention and enhances robustness through representation and data-scale augmentation.
On USPTO-50K, the model achieves strong top-$K$ accuracy and significantly outperforms a vanilla Transformer, demonstrating that template-free approaches can approach the performance of template-based systems when equipped with appropriate structural priors and data augmentation.

\bibliography{reference}

\end{document}